\documentclass[preprint,12pt]{article}
\usepackage[a4paper,margin=1in]{geometry} 
\usepackage[T1]{fontenc}
\usepackage{graphicx}
\usepackage{amsmath, amssymb}
\usepackage{hyperref}
\usepackage{authblk} 
\usepackage[dvipsnames]{xcolor}
\usepackage{tabularx}
\usepackage{booktabs}  
\usepackage{pifont} 
\usepackage{enumitem}
\newcommand{\cmark}{\ding{51}}  
\newcommand{\xmark}{\ding{55}}  

\title{Glioma C6: A Novel Dataset for Training and Benchmarking Cell Segmentation}

\author[1,2]{Roman Malashin}
\author[3]{Svetlana Pashkevich}
\author[1]{Daniil Ilyukhin}
\author[3]{Arseniy Volkov}
\author[1]{Valeria Yachnaya}
\author[3]{Andrey Denisov}
\author[1]{Maria Mikhalkova}

\affil[1]{Pavlov Institute of Physiology, Russian academy of science}
\affil[2]{Saint-Petersburg State University of Aerospace Instrumentation, Russia}
\affil[3]{Institute of Physiology, NAS of Belarus}

\date{}

\begin{document}
\maketitle

\maketitle

\begin{abstract}
  We present Glioma C6, a new open dataset for instance segmentation of glioma
  C6 cells, designed as both a benchmark and a training resource for deep
  learning models. The dataset comprises 75 high-resolution phase-contrast
  microscopy images with over 12,000 annotated cells, providing a realistic
  testbed for biomedical image analysis. It includes soma annotations and
  morphological cell categorization provided by biologists. Additional
  categorization of cells, based on morphology, aims to enhance the utilization
  of image data for cancer cell research. Glioma C6 consists of two parts: the
  first is curated with controlled parameters for benchmarking, while the second
  supports generalization testing under varying conditions. We evaluate the
  performance of several generalist segmentation models, highlighting their
  limitations on our dataset. Our experiments demonstrate that training on
  Glioma C6 significantly enhances segmentation performance, reinforcing its
  value for developing robust and generalizable models. The dataset is publicly
  available for researchers.
\end{abstract}

\section{Introduction}
\label{sec:orgdbee530}

Deep learning has proven to be effective in aiding cancer cell detection in
biopsy-derived tissue samples\cite{kumar2024automating,hesso2023cancer,huhulea2025artificial,jiang2023deep,zhang2025biopsy}, and the primary focus
of our study is \textit{in vitro} tumor cell cultures that offer a complementary
avenue for cancer research\cite{richter2021donor,cirigliano2025bridging,mousavi2021characterization}.

Particularly, we are interested in label-free cell detection and analysis
methods\cite{wang2020label,wu2025antigen,an2025advancements,ahmad2024early}, which provide an alternative to fluorescent
labeling. Fluorescent labeling — a widely adopted technique for cancer cell
detection in microscopy\cite{shen2023automatic} — involves staining cells with
specific fluorescent dyes or genetically encoded markers. This process requires
complex sample preparation and specialized imaging equipment. Moreover, it is
constrained by photobleaching, variability in staining efficiency, and the fact
that it typically necessitates the use of fixed (non-viable) cells, limiting its
applicability for long-term or live-cell studies. In our work, we focus on
phase-contrast microscopy, a technique commonly used for label-free imaging of
\textit{in vitro} cell cultures. However, compared to fluorescence microscopy,
it produces lower-contrast images and introduces specific
artifacts\cite{yin2012understanding}, making image analysis more challenging.
In this context, deep learning plays a crucial role in enabling accurate and
robust analysis.

We also address another key aspect of cancer research and drug
screening\cite{alizadeh2020cellular,basu2014detecting}: the analysis of cell
morphology, commonly approached through phenotyping, i.e. the characterization
of observable features or behaviors. This process includes assessing various
cellular attributes, such as shape, proliferation rates and other phenotypic
markers. Training deep learning models for such analyses requires datasets with
annotated cell shapes, underscoring the need for high-quality labeled data.
While numerous datasets exist
\cite{edlund2021livecell,greenwald2022whole,stringer2021cellpose,cutler2022omnipose,caicedo2019nucleus},
further expansion is necessary to enhance model robustness and generalization.

In this work, we introduce Glioma C6, a novel annotated dataset of
phase-contrast microscopy images of C6 glioma cells, designed to advance cell
segmentation research. Unlike many publicly available datasets, we curate a
specialized dataset with controlled and consistent parameters, enabling reliable
benchmarking in a specialist setting. Additionally, a separate subset of our
dataset allows testing generalization under varying conditions.

The Glioma C6 dataset includes over 12,000 annotated cells in total across 75
high-resolution images, with expert annotations distinguishing between two cell
types. Additionally, soma annotations are provided for 45 images, captured
under consistent shooting conditions.

We summarize our contribution as follows:
\begin{itemize}
\item We collect the first open dataset of glioma C6 cells phase-contrast images
  with soma and cell type annotations.
\item Our experiments demonstrate that generalist models still struggle to
  robustly generalize to new cell datasets without fine-tuning, while fine-tuned
  models deliver reliable and performance, even under varied imaging conditions
  (e.g. variations cultivation time). 
\end{itemize}

The dataset is publicly available for research purposes at
\href{https://zenodo.org/records/15083188}{https://zenodo.org/records/15083188}.

\section{Related works}

In this section, we review the most prominent open datasets used in the
development of cell segmentation frameworks, along with key deep learning
methods for cell segmentation.

\textbf{Datasets}. A variety of datasets have been developed to support the
creation of the universal\cite{stringer2021cellpose} or multimodal cell
segmentation models \cite{ma2024multimodality}; however, many existing
datasets remain limited in diversity, focusing on homogeneous imaging conditions
or cell types. The datasets differ along several dimensions, including image
resolution, cell types, imaging modalities, and acquisition conditions.

Small-scale datasets typically contain from tens to hundreds of images with tens
of thousands of annotated cells. For example, the Cell-APP HeLA dataset consists
of 32 images with around 13500 cells \cite{virdi2025cell}. Other examples
include DSB 2018\cite{caicedo2019nucleus}, which consists of 735 low-resolution
images(up to \(515 \times 512\)) of both live cells and fixed specimens, and
CellPose\cite{stringer2021cellpose}, which includes 551 images.
Dataset \cite{caicedo2019nucleus} was widely used in nuclei detection  \cite{flores2019automate} and segmentation \cite{xiao2023daunet,singha2023alexsegnet,tran2022trans2unet,le2021drunet,manju2023novel} tasks. Medium size
datasets contain several thousands of images. For example,
LiveCell\cite{edlund2021livecell} includes more than 5000 images labeled by
professional annotators under the supervision of cell biologists. This dataset was used in the tasks of the cell center detection \cite{khalid2024cellspot}, cytoplasmic segmentation as well as cell segmentation \cite{raj2024feature,zhou2023scts,khalid2023pace}.
OmniPose\cite{cutler2022omnipose} contains a similar number of images, and only
a subset is annotated. This dataset was also used in the task of the cell segmentation \cite{cazorla2025sketchpose,lou2023multistream}.
 TissueNet is a large-scale dataset
\cite{greenwald2022whole} comprising over 20,000 images of tissue-derived cells
from humans, mice, and macaques. Despite the difference in the number of images,
both LiveCell and TissueNet contain a comparable number of annotated cells, each
exceeding one million.
While several approaches aim to annotate not only cell but also subcellular
features, they usually deal with images obtained with fluorescent or electron
microscopy techniques \cite{chai2024opportunities}.

\textit{Although Glioma C6 is a comparatively small dataset with approximately
  12,000 annotated cells, it stands out due to its unique combination of
  characteristics}.

\textbf{Methods}. In cell segmentation, models designed to generalize across
different cell types and image modalities are referred as generalists, while
those tailored for specific experimental conditions are called
specialists\cite{stringer2021cellpose, stringer2025cellpose3}. One of the first
successful adaption of U-Net \cite{10.1007/978-3-319-24574-4_28} to universal
cell segmentation was CellPose \cite{stringer2021cellpose} which introduced
gradient flows — an output representation that encodes the spatial location of
individual cell instances. The same output format was used in MediarFormer
\cite{lee2022mediar} — winning solution of multimodality cell segmentation
challenge\cite{ma2024multimodality}. MediarFormer incorporated Point-wise
Attention and Multi-scale Fusion Attention Block\cite{fan2020ma}, test-time
augmentation and model ensembles. Recent, CellPose-3
\cite{stringer2025cellpose3} concentrates on enhancing generalist model by
accounting different image degradation processes.

A further step forward is the development of foundational models for medical
image analysis\cite{ma2023towards,lin2024beyond}. For this purpose authors of
\cite{ma2024segment} develop dataset with 1.5m image-mask pairs. They introduced a
promptable model, that allows users to specify segmentation targets using
bounding boxes. The authors of \cite{israel2024foundation} concentrate on developing
foundation model specifically for cell segmentation.

While foundation models represent a promising area, variability in task
definitions and imaging modalities complicates the design of fully automatic
solutions. Pachitariu et al. \cite{pachitariu2022cellpose} advocate
human-in-the-loop tuning of generalist models, demonstrating that fine-tuning
can be label-efficient. More recent
works\cite{archit2025segment,zhou2024cellseg1,lin2024beyond} suggest using
Segment Anything Model (SAM)\cite{kirillov2023segment}.
CellSeg1\cite{zhou2024cellseg1} achieves robust results using only one
labeled image for fine-tuning: authors highlight that annotation quality is more
important than annotation quantity.

\textit{ In this work, we evaluate various approaches, including generalist cell
  segmentation models, under both fine-tuning and zero-shot settings}.

\section{Glioma C6 dataset}

We collect the dataset of phase-contrast images of the C6 glioma cell line,
which is widely utilized in neuro-oncology research, originating from a rat
glial tumor. The C6 cell line is employed to investigate various aspects of
glioma biology, including tumor growth, invasion, and potential therapeutic
interventions\cite{giakoumettis2018c6}. One of the most important phenomena
investigated in cancer cells culture is a proliferation, an increase in the
number of cells as a result of cell growth and cell division (mitosis).

The Cell-APP HeLa dataset \cite{virdi2025cell} is aimed for detection of the
mitosis state, but it is based on detection of chromatin shape on fluorescently
labeled cells. During mitosis, spread cells round up forming a
spherically-shaped volume \cite{cadart2014exploring}. It leads to variations in
phase-contrast imaging artifacts thus complicating algorithmic
analysis\cite{grah2017mathematical}. Variations in cell shape (rounded or
flattened) also allow one to assess the degree of cell adhesion to a specific
culture substrate, estimate influence of specific compounds on the cells state.
Taking into account the importance of the tracking the variations of cells
shape, we propose classifying glioma cells into morphological subtypes and
investigate performance of deep learning methods to detect the subtypes.

\subsection{Methodology}
\label{sec:orgbe0925d}
The process of collecting Glioma C6 dataset included stages of cultivation,
imaging and annotation.

\textbf{Cultivation}. Passaged C6 glial cells (ATCC CCL-107) from the rat Rattus
norvegicus (ATCC CCL-107) were cultured at a concentration of \(2 \times 10^5\)
cells/mL in flasks containing F-12K medium (Kaighn’s modification of Ham’s F-12,
ATCC 30-2004), supplemented with 10\% fetal calf serum (Capricorn, NCS-1A) and a
\(10^{-4}\) g/mL solution of gentamicin sulfate. The flasks were incubated at
37°C in an incubator set to 5\% \(\text{CO}_2\).

\textbf{Imaging}. Images were acquired using a BestScope BS-2092 microscope in
phase-contrast mode with 10× and 20× objective lenses at 24 or 72 hours after
cell suspension seeding. After 24 hours, glioma cells adhered to the Petri dish
surface, acquired a specific shape, and began active division.

\textbf{Overlapped cells} The Glioma C6 dataset differs from typical
segmentation datasets in one important aspect: it includes overlapping
annotations. This was necessary because our goal was to assess the shapes of
individual cells. Therefore, during annotation, we allowed one cell mask to
overlap with another whenever required.

\textbf{Annotation procedure}. For cells with complex morphology, fully manual
segmentation is preferred, as semi-automatic methods often failed to accurately
delineate boundaries. We attempted to speedup the annotation process by
generating initial annotations using Segment Anything
Model\cite{kirillov2023segment}, OpenCV, and pretrained version of
MediarFormer\cite{lee2022mediar}. However, none of these approaches
provided meaningful results in annotation and was not adopted in our procedure.
Subsequently, we employed MediarFormer, fine-tuned on early version of our
dataset, to accelerate soma annotation. The automatically generated results
were then refined through manual correction by the annotators.

The annotations were conducted by a team of biologists experienced in
cell culture and microscopy. There were three human annotators with more than
two years of experience in glioma C6 cells culture. There were several rounds of
annotations; after each round, the results were reviewed by a supervisor, and
corrections were applied until no visible issues remained. However, due to the
frequent occurrence of complex clusters where cells overlap, some unavoidable
uncertainty in the annotation persisted even after careful review. In such
cases, the most probable annotation was chosen.

\subsection{Dataset structure}

We annotate cells, somata, and assign each cell to one of two classes. Overall,
there are three labels in the Glioma C6 dataset:

\begin{enumerate}[label=\Roman*{\Large.}]
\item \textbf{Cell Type A}

  \textit{Description}. These cells correspond to the first phase of growth (early seeding or post-division stage). They are relatively
  loosely attached to the substrate. At this stage, cells exhibit an apparent 3D
  (convex) morphology — their height is comparable to their width and length.

  \textit{Visual cues}. Two main subtypes are observed: spheroid and spindle-shaped.
  \begin{itemize}
  \item Spheroid type A cells appear as small circular structures and typically
 exhibit a distinct high-contrast halo around the cell border in microscope
    images, a feature generally absent in type B cells. They also tend to be
    smaller in area compared to type B.
  \item Spindle-shaped type A cells maintain a distinct 3D form, but are
    elongated along one axis and often show characteristic protrusions.
  \end{itemize}

\item \textbf{Cell Type B}

  \textit{Description}. Type B cells typically correspond to the later phase of
  growth. In this stage, cells firmly attach to the substrate and spread out, so
  their thickness is much smaller than their width and length.

  \textit{Visual cues}. Type B cells are flattened and show lower contrast. When
  viewed from above, they often appear irregularly disk-like, but can also be
  elongated. Their 2D footprint is usually larger than that of type A cells.

\item \textbf{Soma}

  \textit{Description}. Soma contains the nucleus and surrounding cytoplasm, but
  excludes extended protrusions. It is the compact cell body itself,
  representing the functional core of the cell.

  \textit{Visual cues}. Soma includes the central, darker part of the cell, with
  a central region containing a visible nucleus.

\end{enumerate}

The final Glioma C6 dataset consists of two distinct parts:
\begin{itemize}
\item \textbf{Glioma C6-gen} includes 30 images, acquired under varying imaging
  and seeding conditions. We used this dataset to test generalization ability of
  the tuned models.
\item \textbf{Glioma C6-spec} contains 45 images, acquired under strictly
  controlled parameters. We used this dataset to train specialist models for
  controlled experiments.
\end{itemize}

Both subsets feature high resolution $2592 \times 1944$ images, other
characteristics of two parts of dataset are presented in the
\tablename~\ref{tab_glioma_parts}.

\begin{table}[ht]
  \caption{Glioma C6 dataset overview.}
  \label{tab_glioma_parts}
  \centering
  \begin{tabular}{|l|c|c|c|c|c|c|c|}
    \hline
    \textbf{Part} & \textbf{Cells} & \textbf{Soma} & \textbf{Images} & \textbf{Lens} & \textbf{Time, h} & \textbf{Amount} & \textbf{Labels} \\
    \hline
    gen & 3.9K & - &  30 & 10/20 & 24, 72 & Varies & Cell \\
    spec & 8.2K & 7.8K & 45 & 10 & 24 & Same & Cell A/B, soma \\
    \hline
  \end{tabular}
\end{table}

The gen subset was initially used to optimize parameters for the
subsequently collected spec subset, ensuring the most informative imaging
conditions. Preliminary binary segmentation experiments indicated that
preserving image resolution and maintaining consistency in objective lenses
significantly improved segmentation accuracy. We include the gen subset as a
valuable resource for studying the generalization abilities of specialist models
in glioma instance segmentation.

\begin{figure}[ht]
  \centering
  \includegraphics[width=0.6\textwidth]{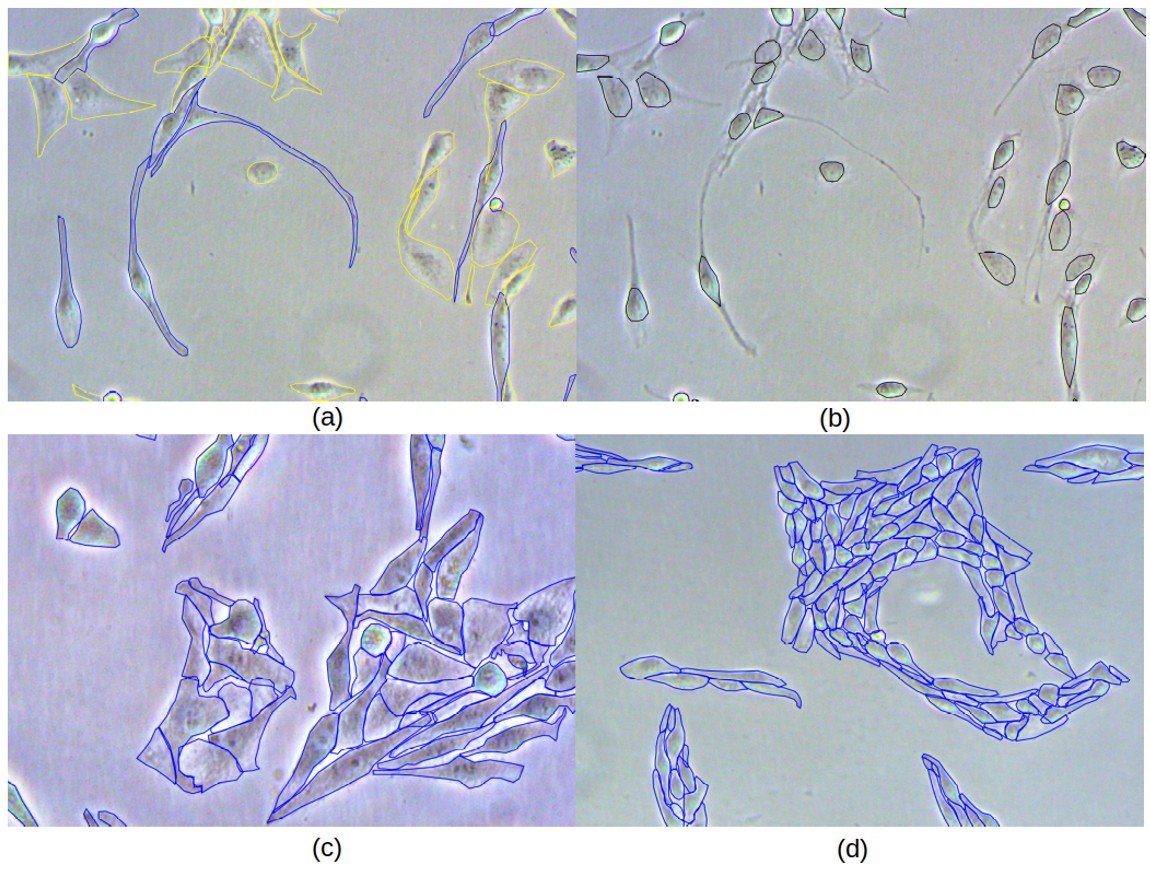}
  \caption{Sample images from the Glioma C6 dataset: (a) \textbf{Type A cells}
    (blue) and \textbf{Type B cells} (yellow), (b) \textbf{Soma}, (c, d)
    \textbf{Cells from gen subset} captured under varying conditions.}
  \label{fig_glioma_examples}
\end{figure}

Visual examples of the Glioma C6 dataset are shown in
\figurename~\ref{fig_glioma_examples}, illustrating different cell types and
structures.

\begin{figure}
  \centering
   \includegraphics[width=0.9\textwidth]{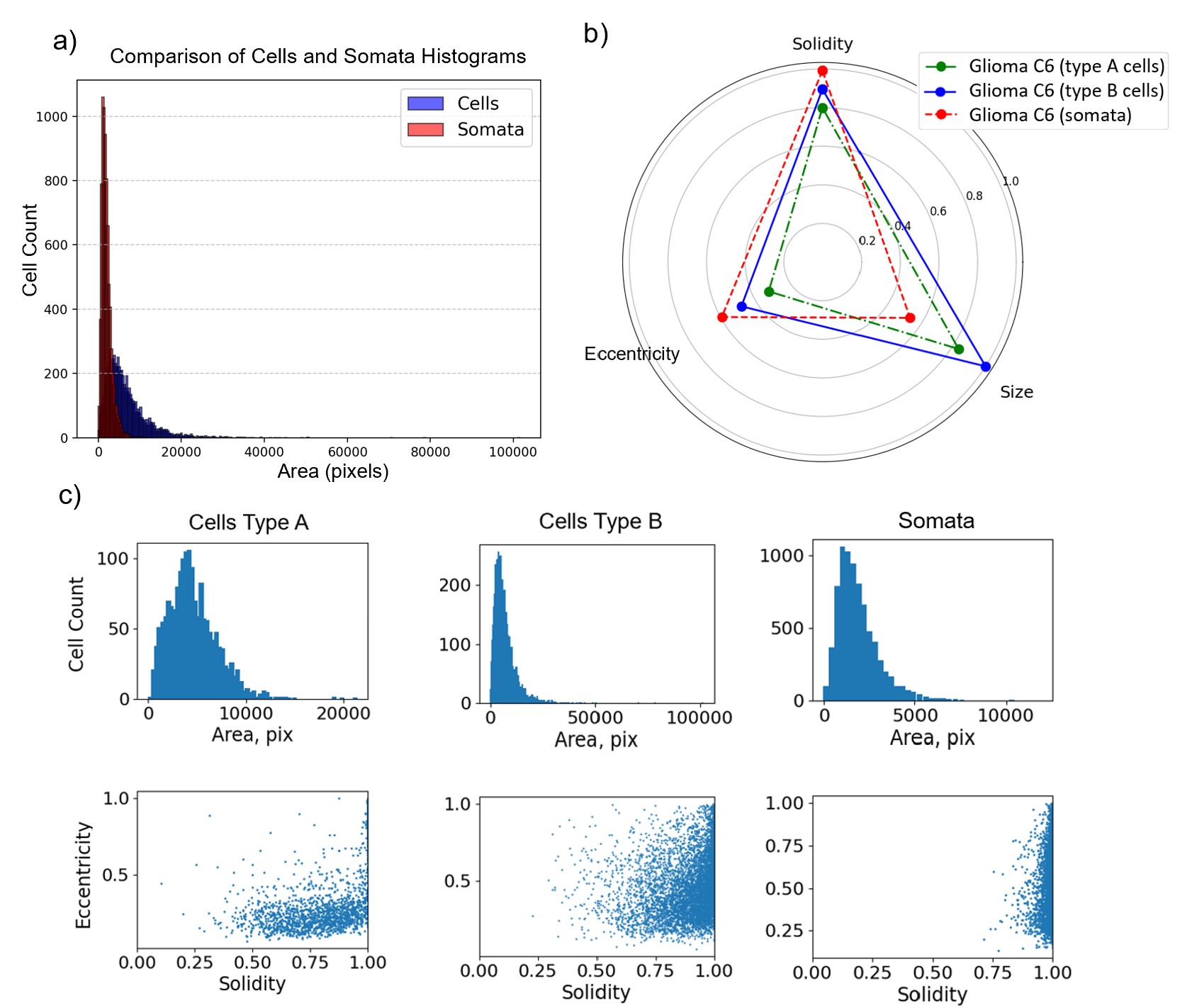}
  \caption{\textbf{Shape distribution}. (a) Cell and soma area distribution
    in spec subset of Glioma C6, (b) Diagram showing eccentricity, solidity and
    size relation of cells type A and type B and soma, (c) Distribution of solidity vs eccentricity and area for glioma C6
    cells and soma.}

    \label{fig:cell_shapes}
\end{figure}

Similarly to\cite{lee2022mediar} we visualize the shape distribution of Type A
and Type B cells through three measures: eccentricity
($\frac{\text{Axis}_{\text{short}}}{\text{Axis}_{\text{long}}}$), solidity
($\frac{\text{Area}}{\text{Convex Area}}$), and cell size
(\figurename~\ref{fig:cell_shapes}).

The figure shows that the mean cell sizes vary across different cell types.
Additionally, distribution of eccentricity vs. solidity ratios exhibits distinct
patterns, particularly between Type A and Type B cells.

In \figurename~\ref{fig:cell_shapes}b we present a diagram visualizing the three
considered measures for both cell types of Glioma C6.

To facilitate comparison, we extracted subsets from well-known cell segmentation
datasets using the same methodology as in \cite{zhou2024cellseg1}:
\begin{itemize}
\item 14 subtypes from TissueNet\cite{greenwald2022whole}.
\item Neuronal cells from CellPose\cite{stringer2021cellpose}.
\item Blood cells from\cite{ma2023towards}.
\item Curated subset from Data Science Bowl (DSB2018) nuclei dataset\cite{caicedo2019nucleus}.
\end{itemize}

\begin{figure}[ht]
  \centering
  \includegraphics[width=0.9\textwidth]{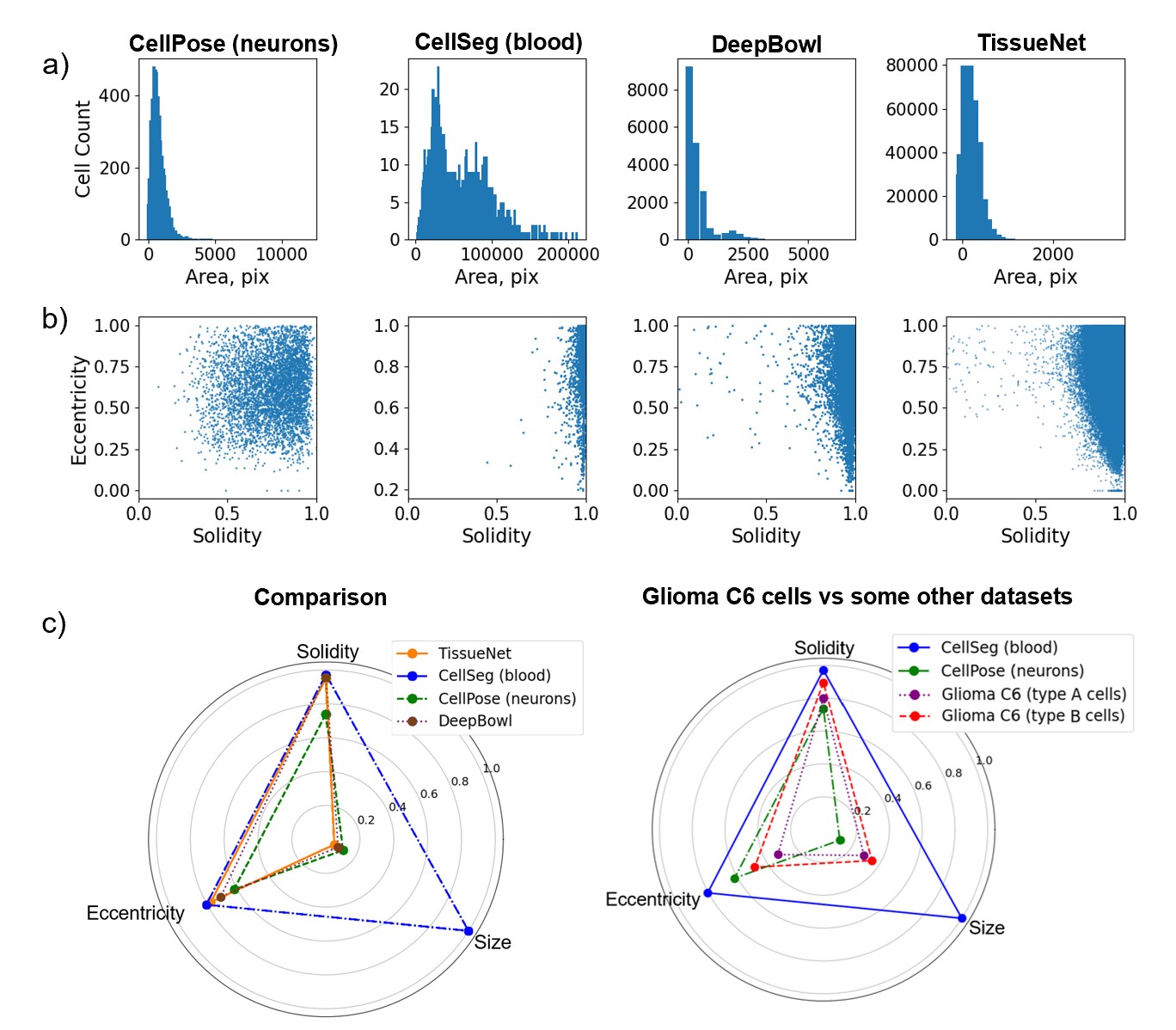}
  \caption{\textbf{Datasets statistics}. (a) Distribution of sizes, (b)
    eccentricity vs. solidity ratio distribution, and (c) Diagram showing
    relation of mean size, solidity and eccentricity of the cell objects in
    different datasets}
  \label{fig:other_datasets}
\end{figure}

The distributions of the characteristics in these datasets are visualized in
\figurename~\ref{fig:other_datasets}.

As can be seen in \figurename~\ref{fig:cell_shapes} and
\figurename~\ref{fig:other_datasets} Type A cells posses unique statistics of
shape.

\section{Experiments}

We selected several models for our main experiments:
YOLOv11\cite{ultralytics2024yolov11}, a widely used general-purpose detector
and instance segmentation model; CellPose \cite{stringer2025cellpose3}, a
well-known generalist model; MediaFormer, an award-winning solution for
multimodality cell segmentation \cite{ma2024multimodality}; and
CellSeg1\cite{zhou2024cellseg1}, a recent low-rank adaptation of the Segment
Anything Model tailored for cell segmentation.

MediarFormer, CellPose, and CellSeg1 are evaluated in both their pretrained form
and after fine-tuning on the Glioma C6 dataset (spec part). In the tables the
\textbf{Glioma-ft} column indicates whether the model was fine-tuned.


Since the cell segmentation models (CellSeg1, MediarFromer, CellPose) cannot be
directly applied to multi-class predictions, we trained separate models for
instance segmentation of cells (both types), soma, cells type~A, cells type~B.
We used only Glioma C6-spec subset for training.

\textbf{Metrics}. To compare model performance, we use the F1 score at an
intersection-over-union (IoU) threshold of 0.5, following
\cite{ma2024multimodality}. Given the inherent uncertainty in manual cell
annotations, an IoU threshold of 0.5 is considered a reasonable criterion for
matching. To ensure a more comprehensive evaluation, we additionally report
precision, recall, and average precision (AP) (similarly to \cite{stringer2021cellpose}) at IoU thresholds of 0.5,
0.75, and 0.9. Consistent with \cite{ma2024multimodality}, cells located at
image boundaries are excluded from all evaluations.

\textbf{Dealing with overlapped cells}. Conventional cell segmentation
algorithms typically assume that individual cells do not overlap. For instance,
diffusion-based approaches employed in CellPose and MediarFormer impose the
constraint that each pixel is assigned to exactly one cell. To enable training
with such models, we converted the Glioma C6 annotations into binary instance
segmentation masks, resolving conflicts deterministically according to the
priority order of the annotated polygons. In contrast, no such conversion was
required for YOLOv11, which inherently supports overlapping instances. During
validation, all performance metrics in the main paper are reported with respect
to the original overlapping annotations. For readers interested in metrics that
disregard overlaps (as in the training sets for MediarFormer and Cellpose), we
report the corresponding results in Appendix ~\ref{appendix:no_overlap}.

Below, we provide details on training the models.
\subsection{Setup}

\textbf{CellSeg1}. We trained CellSeg1 for 300 epochs using the parameters
suggested by CellSeg1 authors for the dataset \cite{ma2024multimodality}, which
includes resolutions comparable to Glioma C6. The default testing pipeline was
suboptimal; however, increasing the number of input crop resolutions — at the
expense of processing speed — significantly improved performance, yielding a gain
of +0.15 AP@0.5. We also observed high variability in the optimal confidence
threshold across training sessions, thus we selected the best value using the
validation set. Training on low-resolution images ($512 \times 512$) used by the
authors can be limiting, but increasing this size (without adjusting other
parameters) significantly degraded performance. Thus, all the results are
reported using these settings.

\textbf{MediarFormer}. The training was conducted for 100 epochs with parameters
proposed by the authors of MediarFormer. Weights from phases 1 and 2 after
fine-tuning on dataset\cite{ma2024multimodality} were used as a starting
checkpoint. Training was performed on $512 \times512$ crops. During testing, crops of
the same size with a 60\% overlap were fed into the network, and the full
resolution masks were recovered prior to metrics compotation. We report results
using an ensemble with Test-Time-Augmentation (TTA), although we did not observe
significant performance gains from their usage.

\textbf{Cellpose}. The CellPose cyto3 model\cite{stringer2025cellpose3} was
trained for 1000 epochs using the hyperparameters recommended by the authors.
The training process incorporated image normalization and augmentation,
including random rotations and scaling by a factor of 0.5. During inference
images were normalized without any further modifications.

\textbf{YOLOv11}. We used the YOLOv11l (large) model with the default parameters
for training and testing. The model was trained for 395 epochs in multi-class
regime (with the use of cell and soma class labels), images for training were
resized to $1024 \times 1024$.  The confidence threshold was kept at the default value of 0.25 during testing.

\subsection{Results}

\begin{figure}[ht]
  \centering
  \includegraphics[width=0.6\textwidth]{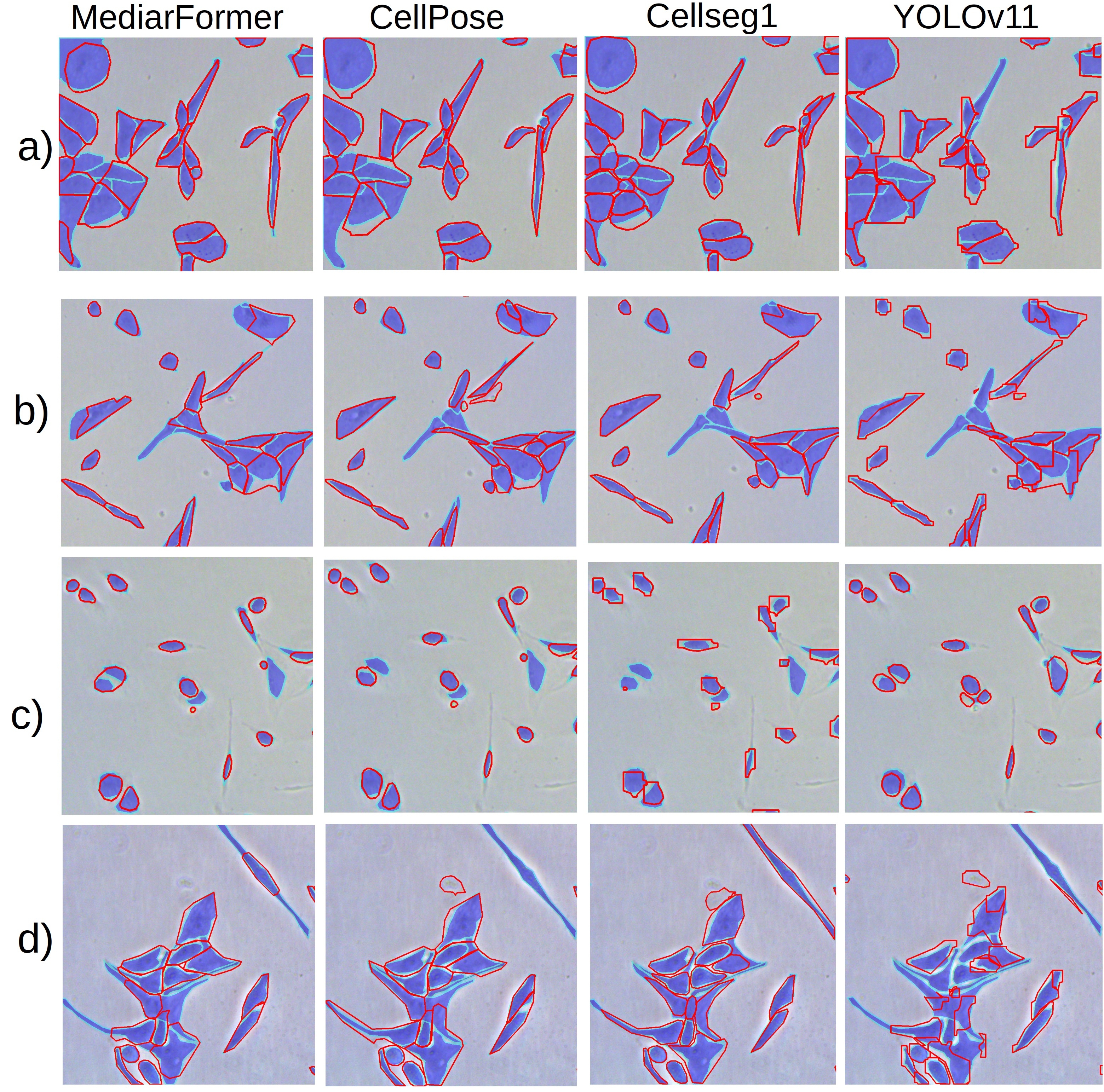}
  \caption{Predictions of the models on the Glioma C6 dataset. Ground truth cell
   borders are shown in light blue, predictions in red, and binary cell
    masks in dark blue. (a, b) Cells, (c) somata. Panels (a), (b)
    and (c) correspond to spec subset, (d) depicts images from the gen subset.}
  \label{fig:visualizations}
\end{figure}

\begin{figure}[ht]
  \centering
  \includegraphics[width=0.6\textwidth]{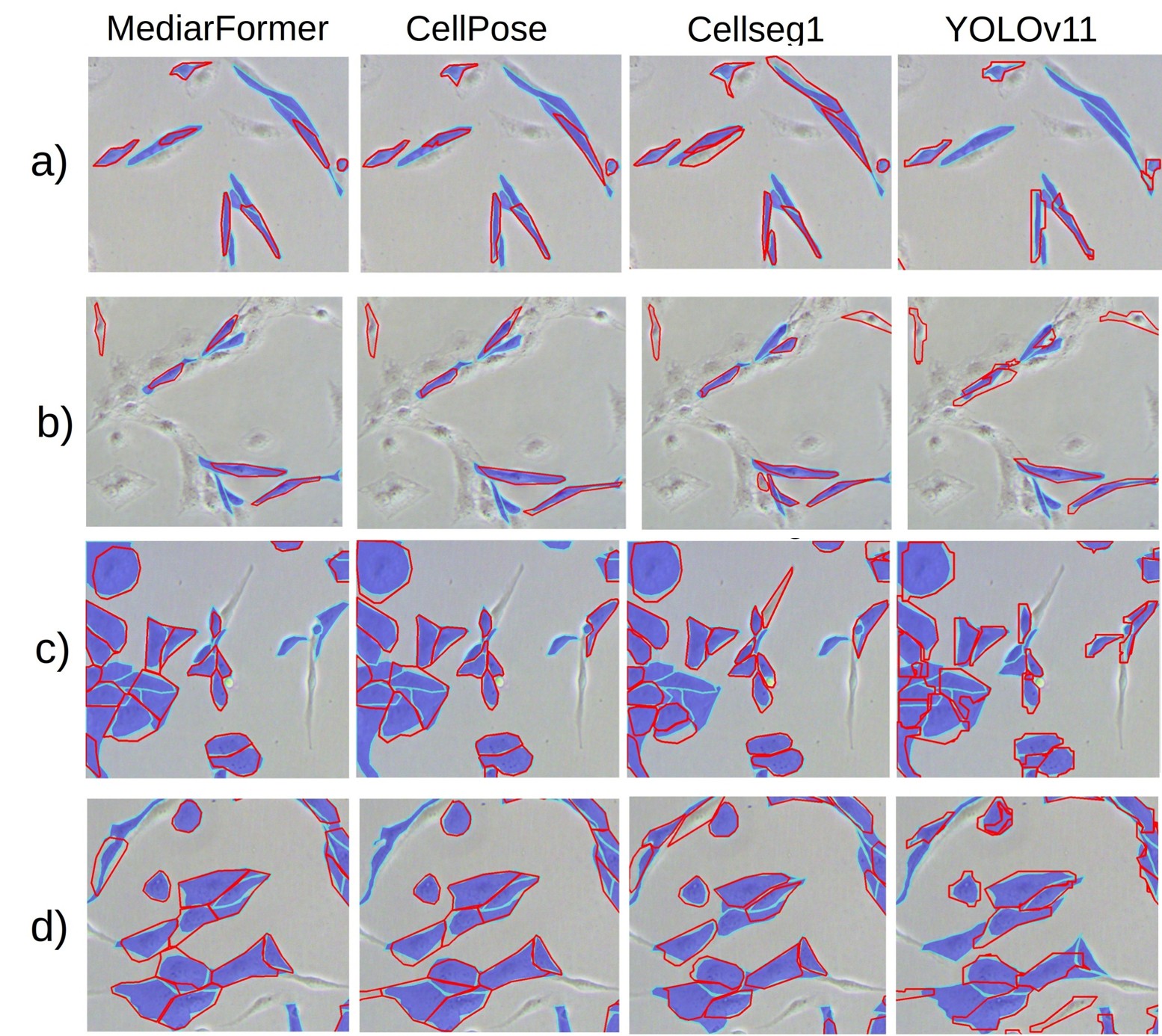}
  \caption{Predictions of the models on the Glioma C6 dataset. (a, b) Type A
    cells, (c, d) Type~B cells. Ground truth cell borders are shown in light
    blue, predictions in red, and binary binary cell masks in dark blue.}
  \label{fig:visualizations_extra}
\end{figure}

We summarize results in \tablename~\ref{tab:results_spec},
\figurename~\ref{fig:visualizations} and
\figurename~\ref{fig:visualizations_extra} provide a visual comparison of the
models' predictions.

\begin{table}[ht]
  \caption{\textbf{Glioma C6-spec performance comparison} of cell segmentation models
    across different categories and subsets.}
    \label{tab:results_spec}
    \centering
    \begin{tabular}{|l|c|c|c|c|c|c|c|}
        \hline
        \textbf{Model} & \textbf{Glioma-ft} & \textbf{Prec.} & \textbf{Rec.} & \textbf{$\mathbf{F_1}$} & \textbf{AP@50} & \textbf{AP@75} & \textbf{AP@90} \\
        \hline
        \multicolumn{8}{|c|}{\textbf{Cells (both types), spec}} \\
        \hline
        \textbf{MediarFormer}  & \color{red}{\xmark} & 57.53 & 2.06 & 3.98 & 2.03 & 0.62 & 0.0 \\
        \textbf{CellPose}      & \color{red}{\xmark} & 21.1 & 11.11 & 14.55 & 7.85 & 1.3 & 0.0 \\
        \hline
        \textbf{YOLOv11}       & \color{ForestGreen}{\cmark} & 70.16 & 78.23 & 73.98 & 58.7 & 11.27 & 0.05 \\
        \textbf{CellSeg1}      & \color{ForestGreen}{\cmark} & 79.15 & \textbf{79.66} & 79.4 & 65.84 & 19.42 & 0.99 \\
        \textbf{MediarFormer}  & \color{ForestGreen}{\cmark} & \textbf{89.32} & 76.86 & 82.62 & 70.39 & 23.45 & \textbf{1.53} \\
        \textbf{CellPose}      & \color{ForestGreen}{\cmark} & 88.87 & 78.87 & \textbf{83.57} & \textbf{71.78} & \textbf{26.31} & 1.29 \\
        \hline
        \multicolumn{8}{|c|}{\textbf{Soma, spec}} \\
        \hline
        \textbf{CellPose}-nuclei  & \color{red}{\xmark} & 0.43 & 0.31 & 0.36 & 0.18 & 0.03 & 0.0 \\
        \hline
        \textbf{YOLOv11}          & \color{ForestGreen}{\cmark} & 77.29 & 77.14 & 77.21 & 62.89 & 4.25 & 0.0 \\
        \textbf{CellSeg1}         & \color{ForestGreen}{\cmark} & 83.52 & 82.15 & 82.83 & 70.69 & 25.0 & 0.49 \\
        \textbf{MediarFormer}     & \color{ForestGreen}{\cmark} & \textbf{89.4} & 85.88 & 87.61 & 77.95 & 32.99 & 0.87 \\
        \textbf{CellPose}         & \color{ForestGreen}{\cmark} & 88.15 & 87.11 & 87.63 & 77.98 & 32.68 & 1.25 \\
        \textbf{CellPose}-nuclei  & \color{ForestGreen}{\cmark} &	87.76	& \textbf{87.76}	& \textbf{87.76} & \textbf{78.19}	&	\textbf{32.87} & \textbf{1.28} \\
        \hline
        \multicolumn{8}{|c|}{\textbf{Cells type A, spec}} \\
        \hline
        \textbf{YOLOv11}      & \color{ForestGreen}{\cmark}  & 71.39 & 56.69 & 63.2 & 46.19 & 15.12 & \textbf{1.2} \\
        \textbf{CellSeg1}     & \color{ForestGreen}{\cmark}  & 44.57 & 75.8 & 56.13 & 39.02 & 13.27 & 0.55 \\
        \textbf{MediarFormer} & \color{ForestGreen}{\cmark}  & 51.41 & 65.61 & 57.65 & 40.5 & 6.67 & 0.0 \\
        \textbf{CellPose}     & \color{ForestGreen}{\cmark}  & \textbf{72.43} & \textbf{65.82} & \textbf{68.97} & \textbf{52.63} & \textbf{18.45} & 1.12 \\
        \hline
         \multicolumn{8}{|c|}{\textbf{Cells type B, spec}} \\
        \hline
        \textbf{YOLOv11}      & \color{ForestGreen}{\cmark}  & 57.04 & \textbf{70.46} & 63.04 & 46.03 & 8.34 & 0.06 \\
        \textbf{CellSeg1}     & \color{ForestGreen}{\cmark}  & 72.31 & 69.31 & 70.78 & 54.78 & 16.77 & 0.99 \\
        \textbf{MediarFormer} & \color{ForestGreen}{\cmark}  & \textbf{85.26} & 59.53 & 70.11 & 53.97 & 16.34 & \textbf{1.45} \\
        \textbf{CellPose}     & \color{ForestGreen}{\cmark}  & 82.19 & 67.26 & \textbf{73.98} & \textbf{58.71} & \textbf{18.85} & 1.14 \\
        \hline
        \multicolumn{8}{|c|}{\textbf{Cells, gen}} \\
        \hline
        \textbf{MediarFormer}  & \color{red}{\xmark} & \textbf{85.43} & 20.92 & 33.6 & 20.19 & 9.06 & 1.01 \\
        \textbf{CellPose}      & \color{red}{\xmark} & 30.81 & 29.08 & 29.92 & 17.59 & 5.26 & 0.41 \\
        \hline
        \textbf{YOLOv11}       & \color{ForestGreen}{\cmark}  & 53.89 & 69.96 & 60.88 & 43.77 & 4.71 & 0.03 \\
        \textbf{CellSeg1}      & \color{ForestGreen}{\cmark}  & 66.17 & \textbf{81.08} & 72.87 & 57.32 & 23.36 & \textbf{2.06} \\
      \textbf{MediarFormer}  & \color{ForestGreen}{\cmark}  & 83.29 & 77.64 & 80.41 & 67.24 & \textbf{26.4} & 1.45 \\
      \textbf{CellPose}      & \color{ForestGreen}{\cmark}  & 82.63 & 79.08 & \textbf{80.82} & \textbf{67.81} & 24.78 & 1.0 \\
        \hline
    \end{tabular}
\end{table}

As expected, YOLOv11 performed poorly, reinforcing the notion that it is not
well-suited for cell segmentation tasks. CellSeg1 demonstrates competitive
recall, but has significantly lower precision compared to other cell
segmentation methods, resulting in reduced overall performance as measured by
F1 and AP.

Overall, CellPose achieved the highest performance in both specialist and
generalization testing, slightly outperforming MediarFormer. While MediarFormer
exhibited higher precision, CellPose attained superior recall. Although
performance decreased for all models on the gen subset, it remained
stable and sufficiently high across diverse imaging conditions.

\figurename~\ref{fig:visualizations_extra} demonstrates 
models' predictions on the Glioma C6 dataset for type A and type B cells.

\section{Assessment of the annotation uncertainty}
\label{sec:uncertainty}

Some annotation decisions have inherently unavoidable uncertainty due to the
complexity of the annotating clusters of cells. This uncertainty cannot be
resolved by human experts relying on the available image data. That means that
the same expert can make different annotation decisions based on the same image
data. This suggests that the dataset contains inaccuracies, which should limit
the maximum AP the ideal algorithm can obtain on Glioma C6 dataset.

In this study, we aim to quantify this uncertainty by performing an independent
re-annotation of part of the Glioma C6 dataset. Three experts each independently
annotated two images from the Glioma C6 test set. The first image (image~1) was
selected as the image with the highest AP@0.5 according to MediarFormer
predictions on the original dataset annotation, and the second image (image~2)
was selected as the image with the lowest score, where we expect annotation to
be particularly challenging. After completing their independent annotations, the
experts engaged in a joint review to produce a consensus (consolidated
annotation). This consensus annotation was then treated as the reference for
comparison against (i) the original Glioma C6 annotation and (ii) CellPose
predictions. The comparison results are summarized in
\tablename~\ref{tab:expert_results}.

\begin{table}[ht]
  \caption{ Assessment of annotation uncertainty in the Glioma C6 dataset via
    comparison with independent expert annotations. Three experts independently
    re-annotated two test images, then produced a consensus annotation through
    discussion. The table reports the agreement between different annotation
    sources (original dataset annotation and CellPose predictions) and the
    expert consensus.}
  \label{tab:expert_results}
  \centering
  \begin{tabular}{|l|c|c|c|c|c|c|}
    \hline
    \textbf{Annotation} & \multicolumn{2}{c|}{\textbf{AP@50}} & \multicolumn{2}{c|}{\textbf{AP@75}} & \multicolumn{2}{c|}{\textbf{AP@90}} \\
    \cline{2-7}
    & image 1 & image 2 & image 1 & image 2 & image 1 & image 2 \\
    \hline
    \multicolumn{7}{|c|}{\textbf{Cells (both types)}} \\
    \hline
    Dataset annotation     & 0.7137 & 0.8117 & 0.1394 & 0.3478 & 0.0000 & 0.0072 \\
    CellPose               & 0.7445 & 0.8151 & 0.2073 & 0.4171 & 0.0051 & 0.0232 \\
    \hline
    \multicolumn{7}{|c|}{\textbf{Soma}} \\
    \hline
    Dataset annotation     & 0.7792 & 0.8514 & 0.2607 & 0.2925 & 0.0049 & 0.0000 \\
    CellPose               & 0.8739 & 0.8562 & 0.3419 & 0.3897 & 0.0146 & 0.0226 \\
    \hline
    \multicolumn{7}{|c|}{\textbf{Cells type A}} \\
    \hline
    Dataset annotation     & 0.6667 & 0.6744 & 0.0588 & 0.3091 & 0.0000 & 0.0000 \\
    CellPose               & 0.6170 & 0.7297 & 0.0704 & 0.3913 & 0.0000 & 0.0000 \\
    \hline
    \multicolumn{7}{|c|}{\textbf{Cells type B}} \\
    \hline
    Dataset annotation     & 0.6262 & 0.7395 & 0.1512 & 0.3185 & 0.0000 & 0.0098 \\
    CellPose               & 0.6062 & 0.7477 & 0.1969 & 0.2897 & 0.0065 & 0.0331 \\
    \hline
  \end{tabular}
\end{table}

\begin{figure}[ht]
  \centering
  \includegraphics[width=0.6\textwidth]{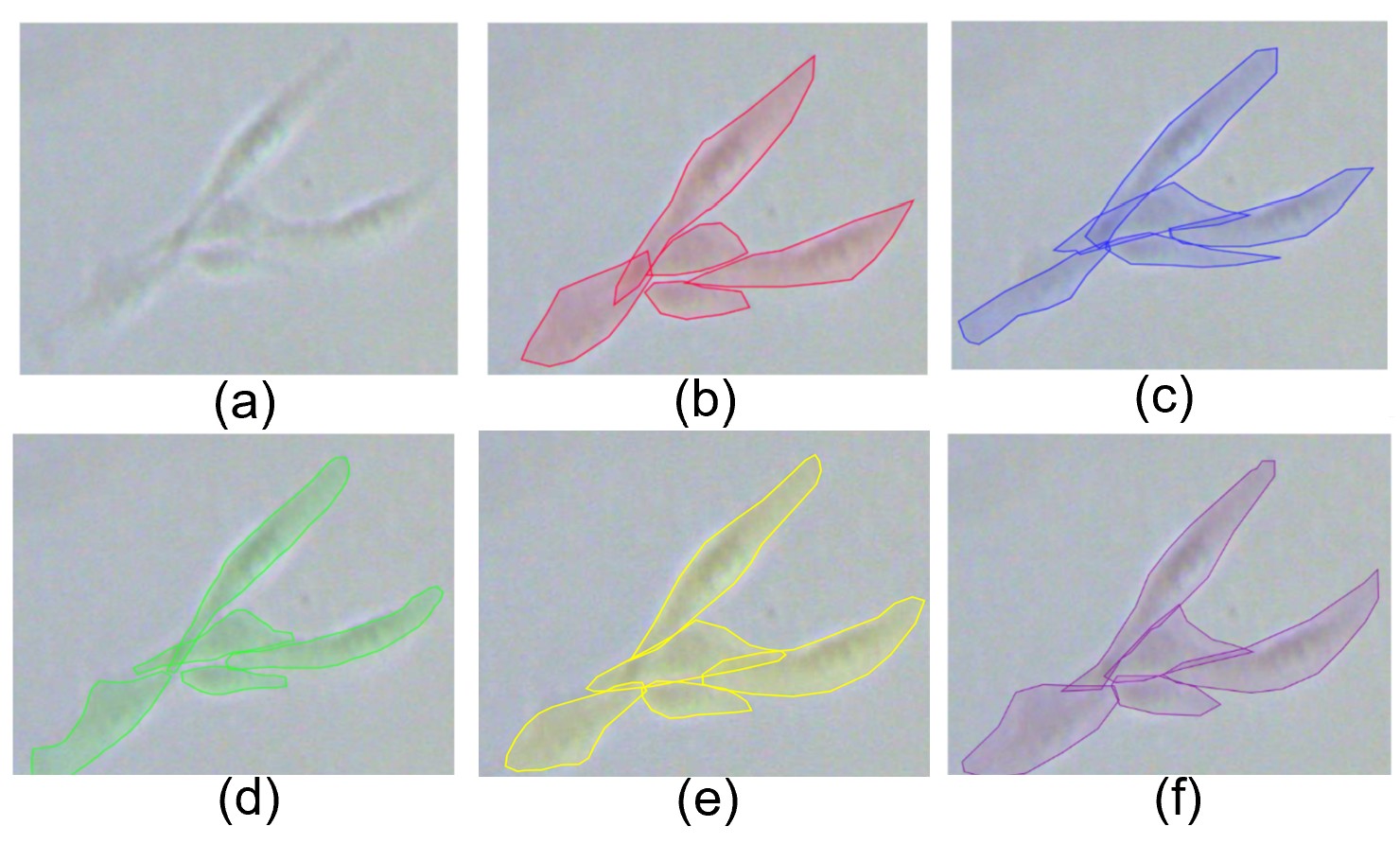}
  \caption{Differences in cell annotations by experts: (a) original image, (b)
    annotation by the expert-1, (c) annotation by the expert-2, (d) annotation
    by the expert-3, (e) consolidated annotation by the experts, (f) Glioma C6
    dataset annotation. }
  \label{fig:experts}
\end{figure}

The results indicate that the most informative evaluation threshold is AP@0.5.
At higher IoU thresholds such as 0.75 and 0.90, the reported values decrease
sharply and in many cases approach zero. This behavior highlights how difficult
it is, even for expert annotators, to achieve pixel-accurate agreement on cell
boundaries in crowded and low-contrast regions. The qualitative analysis in
\figurename~\ref{fig:experts} confirms that, in challenging regions, experts can
legitimately disagree on where one cell ends and another begins. Importantly,
these discrepancies typically represent plausible alternative segmentations
rather than clear annotation mistakes.

Furthermore, the subtype-specific scores in \tablename~\ref{tab:expert_results}
(cell type~A vs.\ cell type~B) show that fine-grained classification of cell
subtypes is particularly challenging. The metrics for individual subtypes are
consistently lower than those for the combined cell class or for somata. This
suggests that distinguishing between phenotypically similar subtypes is less
reliable than detecting “cell vs.\ background” in general.

The dataset annotation itself shows imperfect agreement with the expert
consensus, which indicates that the provided ground truth is not fully
self-consistent. This gap likely reflects both the inherent difficulty of
tracing boundaries in tightly packed cell clusters and potential annotation
fatigue or ambiguity during the original labeling process. Consistent with this
interpretation, the Glioma C6 annotation for the image associated with lower
MediarFormer scores also exhibits weaker agreement with the expert consensus
than the annotation for the image associated with higher MediarFormer scores.

Interestingly, CellPose predictions achieve higher AP scores than the original
Glioma C6 annotations when both are evaluated against the expert consensus. This
suggests that the model has learned an internally consistent annotation policy
and can generate segmentations that are, at least in some cases, more regular
and less noisy than the original ground truth. In other words, model training
can act as an implicit denoising step: it reduces annotation ambiguity and
enforces consistent boundary placement across samples.

Despite these limitations, the Glioma C6 dataset remains highly valuable: it
captures realistic, densely packed cellular environments with complex morphology
and overlapping structures, i.e. exactly the regimes where automated
segmentation is most clinically and biologically relevant. In this sense, the
observed disagreement is not a flaw of the dataset itself, but a reflection of
the genuine ambiguity of the underlying biological signal.

\section{Conclusions}

We believe the Glioma C6 dataset serves as a valuable resource for advancing in
segmenting and quantifying individual cells within developing dense tumor
microenvironments. Accurate instance segmentation of glioma cells enables
precise characterization of cell morphology, proliferation patterns, and
responses to therapeutic interventions. The dataset introduces unique features
that enhance its applicability in specialized studies. In particular, the
introduction of two classes of cells based on qualitative features aims to
explore the potential of deep learning methods for cell phenotyping, including
the potential to detect subtle subcellular variations that are difficult to
analyze algorithmically. The results indicate that generalist models exhibit
limited generalization to Glioma C6 test images, while fine-tuned models achieve
consistent and accurate performance under varied imaging and culturing
conditions. This highlights the crucial role of the collected dataset in
facilitating model adaptation for glioma segmentation.

\section{Acknowledgments}
The work was carried out with the financial support of a grant from the St.
Petersburg Science Foundation. The study was supported by the State funding
allocated to the Pavlov Institute of Physiology, Russian Academy of Sciences
(№1021062411653-4-3.1.8). This work was financially supported by BRFFR (SCST)
with the grant “Detection of tumor cells in nervous tissue using deep learning
methods” (contract No. M24SPbG-010).

\appendix

\section{Additional metrics without overlap}
\label{appendix:no_overlap}

In \tablename~\ref{tab:results_spec}, the metrics were computed under the
assumption that object instances may overlap. By contrast, the evaluation in
\tablename~\ref{tab:results_spec_old} was performed on non-overlapping masks,
where all conflicts between instances were resolved deterministically using the
same procedure that was applied during training for both MediarFormer and
CellPose.

\begin{table}[ht]
  \caption{\textbf{Glioma C6-spec performance comparison} of cell segmentation models
    across different categories and subsets without overlapping.}
    \label{tab:results_spec_old}
    \centering
    \begin{tabular}{|l|c|c|c|c|c|c|c|}
        \hline
        \textbf{Model} & \textbf{Glioma-ft} & \textbf{Prec.} & \textbf{Rec.} & \textbf{$\mathbf{F_1}$} & \textbf{AP@50} & \textbf{AP@75} & \textbf{AP@90} \\
        \hline
        \multicolumn{8}{|c|}{\textbf{Cells  (both types), spec}} \\
        \hline
        \textbf{MediarFormer}  & \color{red}{\xmark} & 56.0 & 2.2 & 4.2 & 2.1 & 0.7 & 0.0 \\
        \textbf{CellPose}      & \color{red}{\xmark} & 19.8 & 11.4 & 14.5 & 7.8 & 1.4 & 0.0 \\
        \hline
        \textbf{YOLOv11}       & \color{ForestGreen}{\cmark} & 55.1 & 70.3 & 61.8 & 44.7 & 6.1 & 0.0 \\
        \textbf{CellSeg1}      & \color{ForestGreen}{\cmark} & 74.9 & \textbf{78.1} & 76.5 & 61.9 & 19.2 & 0.9 \\
        \textbf{MediarFormer}  & \color{ForestGreen}{\cmark} & \textbf{86.8} & 74.4 & 80.1 & 66.8 & 22.0 & \textbf{1.2} \\
        \textbf{CellPose}      & \color{ForestGreen}{\cmark} & 86.5 & 77.2 & \textbf{81.6} & \textbf{69.0} & \textbf{25.4} & 1.0 \\
        \hline
        \multicolumn{8}{|c|}{\textbf{Soma, spec}} \\
        \hline
        \textbf{CellPose}-nuclei  & \color{red}{\xmark} & 0.5 & 0.4 & 0.4 & 0.2 & 0.0 & 0.0 \\
        \hline
        \textbf{YOLOv11}          & \color{ForestGreen}{\cmark} & 74.0 & 76.0 & 75.0 & 60.0 & 4.3 & 0.0 \\
        \textbf{CellSeg1}         & \color{ForestGreen}{\cmark} & 82.8 & 81.1 & 81.9 & 69.4 & 22.2 & 0.3 \\
        \textbf{MediarFormer}     & \color{ForestGreen}{\cmark} & \textbf{89.3} & 84.6 & 86.9 & 76.9 & 32.8 & 0.7 \\
        \textbf{CellPose}         & \color{ForestGreen}{\cmark} & 89.0 & 86.9 & \textbf{87.9} & \textbf{78.5} & \textbf{33.9} & 1.4 \\
        \textbf{CellPose}-nuclei  & \color{ForestGreen}{\cmark} &	88.0	& \textbf{87.7}	& 87.8 & 78.3	&	33.2 & \textbf{1.5} \\
        \hline
        \multicolumn{8}{|c|}{\textbf{Cells type A, spec}} \\
        \hline
        \textbf{YOLOv11}      & \color{ForestGreen}{\cmark}  & 37.7 & 55.5 & 44.9 & 28.9 & 4.6 & 0.0 \\
        \textbf{CellSeg1}     & \color{ForestGreen}{\cmark}  & 44.3 & \textbf{73.2} & 55.2 & 38.1 & 14.4 & 0.8 \\
        \textbf{MediarFormer} & \color{ForestGreen}{\cmark}  & \textbf{70.8} & 57.7 & 63.5 & 46.6 & 16.2 & 1.0 \\
        \textbf{CellPose}     & \color{ForestGreen}{\cmark}  & 65.6 & 66.7 & \textbf{66.1} & \textbf{49.4} & \textbf{19.1} & \textbf{1.2} \\
        \hline
        \multicolumn{8}{|c|}{\textbf{Cells type B, spec}} \\
        \hline
        \textbf{YOLOv11}      & \color{ForestGreen}{\cmark}  & 42.9 & 59.1 & 49.7 & 33.1 & 5.2 & 0.1 \\
        \textbf{CellSeg1}     & \color{ForestGreen}{\cmark}  & 69.0 & \textbf{68.5} & 68.8 & 52.4 & 16.3 & \textbf{1.0} \\
        \textbf{MediarFormer} & \color{ForestGreen}{\cmark}  & \textbf{81.5} & 57.7 & 67.6 & 51.0 & 15.2 & 1.0 \\
        \textbf{CellPose}     & \color{ForestGreen}{\cmark}  & 78.7 & 65.9 & \textbf{71.7} & \textbf{55.9} & \textbf{17.7} & 0.9 \\
        \hline
        \multicolumn{8}{|c|}{\textbf{Cells, gen}} \\
        \hline
        \textbf{MediarFormer}  & \color{red}{\xmark} & 81.2 & 20.4 & 32.6 & 19.5 & 8.3 & 0.7 \\
        \textbf{CellPose}      & \color{red}{\xmark} & 30.1 & 31.6 & 30.8 & 18.2 & 5.5 & 0.5 \\
        \hline
        \textbf{YOLOv11}       & \color{ForestGreen}{\cmark}  & 38.2 & 59.0 & 46.4 & 30.2 & 2.7 & 0.0 \\
        \textbf{CellSeg1}      & \color{ForestGreen}{\cmark}  & 62.7 & \textbf{81.1} & 70.7 & 54.7 & 21.1 & \textbf{1.6} \\
      \textbf{MediarFormer}  & \color{ForestGreen}{\cmark}  & \textbf{80.4} & 75.4 & 77.8 & 63.7 & \textbf{21.2} & 1.0 \\
      \textbf{CellPose}      & \color{ForestGreen}{\cmark}  & 80.3 & 76.8 & \textbf{78.5} & \textbf{64.6} & 19.9 & 0.7 \\
        \hline
    \end{tabular}
\end{table}
As can be seen from the table, all conclusions drawn from
\tablename~\ref{tab:results_spec} remain valid. Notably, the lower metrics reported
in \tablename~\ref{tab:results_spec_old} indicate potential for further improvement
in MediarFormer and CellPose, particularly if their architectures are modified
to accommodate overlapping cells.

\end{document}